\documentclass[conference]{IEEEtran}
\IEEEoverridecommandlockouts

\usepackage{cite}
\usepackage{amsmath,amssymb,amsfonts}
\usepackage{algorithmic}
\usepackage{graphicx}
\usepackage{textcomp}
\usepackage{xcolor}

\usepackage{url}
\usepackage{booktabs}       
\usepackage{amsfonts}       
\usepackage{nicefrac}       
\usepackage{microtype}      

\usepackage{multirow}
\usepackage{array}
\usepackage{verbatim}
\usepackage{caption}
\usepackage{indentfirst}
\usepackage{setspace}
\usepackage{algorithm, listings}
\usepackage{mathtools}
\usepackage{commath}
\usepackage{hhline}
\usepackage{subfigure}

\usepackage{subcaption}
\usepackage{wrapfig}
\usepackage{float}
\usepackage{enumitem}
\usepackage{bm}

\def\BibTeX{{\rm B\kern-.05em{\sc i\kern-.025em b}\kern-.08em
    T\kern-.1667em\lower.7ex\hbox{E}\kern-.125emX}}

\begin{document}

\title{

AgentPose: Progressive Distribution Alignment via Feature Agent for Human Pose Distillation \\

\thanks{Lei Chen is the corresponding author.}

\author{
\IEEEauthorblockN{
 Feng Zhang\IEEEauthorrefmark{2},
 Jinwei Liu\IEEEauthorrefmark{2},
 Xiatian Zhu\IEEEauthorrefmark{3},
 Lei Chen\IEEEauthorrefmark{2}}
\IEEEauthorblockA{\IEEEauthorrefmark{2}
\textit{Nanjing University of Posts and Telecommunications, Nanjing, China} \\
\{zhangfengwcy,~jinweiliu89\}@gmail.com, chenlei@njupt.edu.cn}
\IEEEauthorblockA{\IEEEauthorrefmark{3}
\textit{Surrey Institute for People-Centred 
Artificial Intelligence, University of Surrey Guildford, United Kingdom} \\
eddy.zhuxt@gmail.com}
}
}

\maketitle

\begin{abstract}

Pose distillation is widely adopted to reduce model size in human pose estimation. However, existing methods primarily emphasize the transfer of teacher knowledge while often neglecting the performance degradation resulted from the curse of capacity gap between teacher and student. To address this issue, we propose AgentPose, a novel pose distillation method that integrates a feature agent to model the distribution of teacher features and progressively aligns the distribution of student features with that of the teacher feature, effectively overcoming the capacity gap and enhancing the ability of knowledge transfer. Our comprehensive experiments conducted on the COCO dataset substantiate the effectiveness of our method in knowledge transfer, particularly in scenarios with a high capacity gap.
\end{abstract}


\section{Introduction}
Currently, the growing demand for real-time performance has driven researchers to prioritize the development of lightweight models for human pose estimation\cite{zheng2023deep, chen20232d,stadler2021improving}.  
Previous efforts\cite{ye2023distilpose,yang2023effective,li2021online}  widely adopt pose distillation to transfer pose-related knowledge between models of varying capacities, aiming to achieve a trade-off between efficiency and accuracy.
However, many researchers recognize the problem of capacity gap\cite{mirzadeh2020improved,qian2022switchable,ding2023skdbert}, that is, the more powerful teacher may not always yield a better-performing student and even harm the student’s performance.

To mitigate the capacity gap curse in human pose estimation, we propose AgentPose, a novel diffusion-based knowledge distillation framework.
This approach harnesses the generative capabilities of diffusion models\cite{DBLP:conf/nips/HoJA20,Rombach_2022_CVPR} and incorporates a lightweight feature agent, which facilitates knowledge transfer between teacher and student models, effectively bridging the capacity gap.
Specifically, the feature agent is equipped with feature distribution perturbation and dynamic distribution modulation.
The feature distribution perturbation is tailored to direct distribution of student and teacher features towards an intermediate state that facilitates a smoother process of knowledge transfer.
Subsequently, we introduce appropriate Gaussian noise into the student feature to hijack the reverse SDE (Stochastic Differential Equation) process. This enables the dynamic adjustment of student feature distributions, minimizing the distribution discrepancies between the teacher and student to the greatest extent.
This strategic adjustment of distribution enhances the consistency of learning preference between models, thereby facilitating a more conducive environment for knowledge transfer.

In addition, to reduce the computational load, we introduced an autoencoder\cite{Rombach_2022_CVPR} to diminish the dimensionality of the features before inputting them into the agent module.
Comprehensive experiments demonstrate that our framework exceeds existing pose distillation methods, particularly in scenarios with significant capacity gaps between the teacher and the student. 

Our contributions are summarized as follows:
\begin{itemize}
\item We propose AgentPose, a pose distillation framework that integrates a feature agent for human pose estimation. 
This framework adeptly aligns student knowledge with teacher knowledge by dynamically modulating feature distributions, effectively bridging the capacity gap that always plagues pose distillation.
\item  We further accelerate model inference through constructing a lightweight model architecture and reducing the dimension of features processed by the feature agent with an autoencoder.
\item Extensive experiments demonstrate that AgentPose facilitates efficient knowledge transfer, enabling the compact model to achieve superior performance, particularly in scenarios where there is a huge capacity gap between student and teacher.
\end{itemize}

\section{Method}

\begin{figure*}[ht]
\centerline{\includegraphics[width=0.9\textwidth]{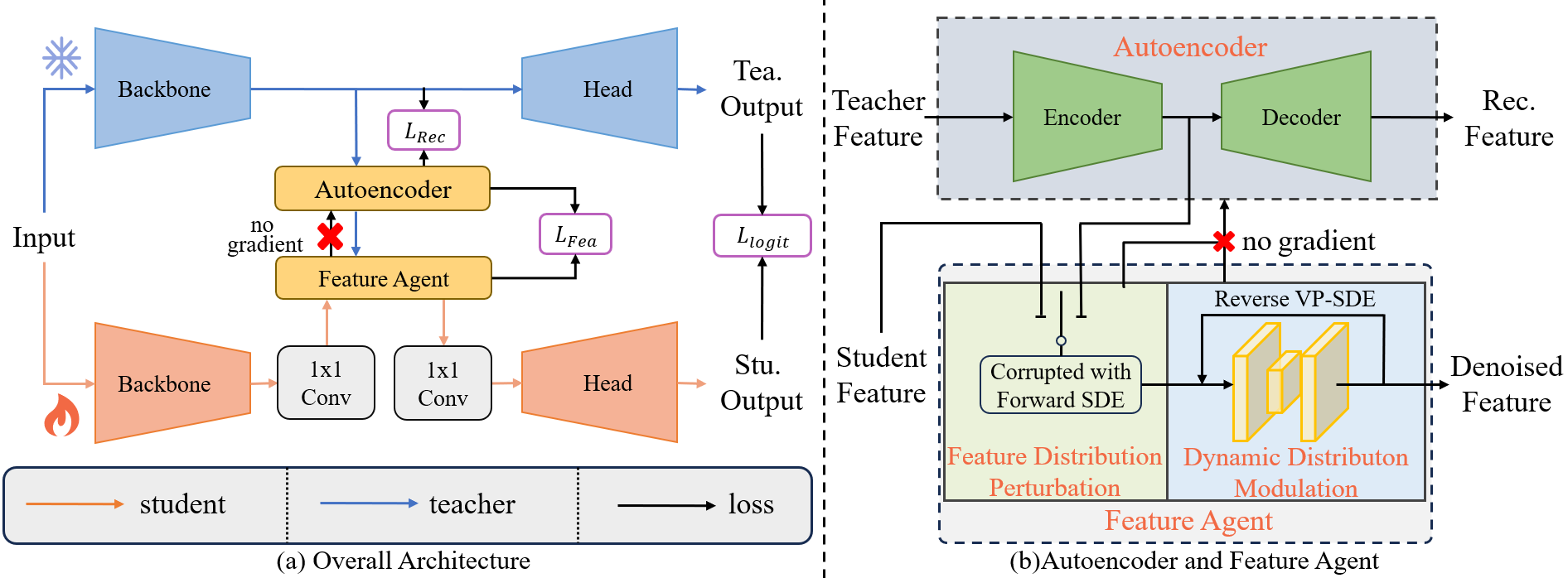}}
\caption{The overview of AgentPose. (a) The architecture of AgentPose, (b) Autoencoder and Feature Agent. Feature agent is trained using corrupted teacher feature, and utilizes a specific reverse VP-SDE (variance preserving stochastic differential equation) to calibrate student feature to enhance the effectiveness of pose distillation. Furthermore, an autoencoder and two convolution layers are included in the AgentPose to reduce the computational overhead of feature agent.}

\label{fig:model}
\vspace{-2em}
\end{figure*}

As depicted in Fig.\ref{fig:model}, AgentPose extends a basic pose distillation framework by integrating the proposed feature agent.
This module harnesses the generative power of the diffusion model to dynamically modulate the student feature, thereby reducing the feature discrepancies that arise from the capacity gap between the teacher and student models.

\subsection{Feature Agent}

The proposed feature agent comprises feature distribution perturbation and dynamic feature distribution modulation. 
The former controls the intensity of noise added to teacher and student features,
while the latter uses a diffusion model to transform noise into features, enabling the learning of the distribution of teacher features.
In addition, since adding noise can help align distributions of student and teacher features towards a similar intermediate state, the feature agent further applies noise perturbation to the student feature via the feature distribution perturbation, allowing dynamic feature distribution modulation to calibrate the student knowledge.

\subsubsection{Feature Distribution Perturbation}

The feature distribution perturbation of feature agent applies noise with different intensities to the teacher feature $F_{tea}$ and student feature $F_{stu}$. 
For the sake of simplicity, We denote the features of both the student and the teacher as $x$ and employ forward Variance Preserving SDE (VP-SDE) to introduce perturbations:
    \begin{align}
               dx(t) &= - \frac{1}{2} \beta(t)x(t)dt + \sqrt{\beta(t)}dw  \label{noise} \\
               \beta(t) &= \beta_{min} + t(\beta_{max} - \beta_{min})
    \end{align}
where $x(t)$ is features with varying degrees of noise, $t \in [0,1]$ denotes the timestep of the forward VP-SDE and controls the intensity of noise. When $t=0$, $x(0)$ represents $F_{tea}$ or $F_{stu}$ without any noise. The random process $w$ is determined by the Wiener process $z(t) \sim N(0,I)$, which satisfies $z(t) = w(t)-w(t-\Delta t)$, with $\Delta t = 1/N$ being the step size of $N$-step forward VP-SDE. $\beta(t)$ is a positive function that controls the speed of noise diffusion. Following previous efforts\cite{DBLP:conf/nips/HoJA20, song2021scorebased}, we set $\beta_{min}=0.0001$ and $\beta_{max}=0.02$ to ensure $p_{t=1}(x) \approx N(0,I)$.

Both $F_{tea}$ and $F_{stu}$ adopt forward VP-SDE to introduce perturbations, but the choice of timestep $t$ depends on whether features come from the teacher or student.
For features from the teacher, $t$ takes random values on $[0, 1]$ to generate various noisy samples.  
These noisy samples enable the feature agent to recover posture information, ensuring that the resulting distribution aligns closely with that of the teacher's features.
In contrast, for features from the student, we set $t$ to a fixed value $t_s = 0.4$ to smooth out unexpected posture information that is inconsistent with the teacher.

\begin{table*}[ht]
    \centering
    \caption{Comparison with the foundational pose estimation method RTMPose and other pose distillation methods in human pose estimation. The values in brackets represent the performance distinction between the method and corresponding RTMPose with the same backbone. RTMPose does not use the distillation strategy as a baseline. our proposed AgentPose employs RTMPose-L as the teacher model, while the smaller RTMPose variants (S/M/T) serve as student models.
    }
    \resizebox{0.95\linewidth}{!}{
  \begin{tabular}{|c|c|c|c|c| c| c |c |c |c|}
    \hline
        \textbf{Method} & \textbf{Teacher} & \textbf{Student} &  \textbf{GFLOPs} & $\bm{AP}$ & $\bm{AP^{50}}$ & $\bm{AP^{75}}$ & $\bm{AP^M}$ & $\bm{AP^L}$ & $\bm{AR}$  \\ \hline
        RTMPose\cite{jiang2023rtmpose} & - & RTMPose-M &  1.9 & 74.3 & 90.0 & 81.3 & 70.7 & 81.0 & 79.3  \\ 
        RTMPose\cite{jiang2023rtmpose} & - & RTMPose-S &  0.7 & 71.9 & 89.4 & 79.4 & 68.3 & 78.4 & 77.0  \\ 
        RTMPose\cite{jiang2023rtmpose} & - & RTMPose-T &  0.4 & 67.9 & 88.1 & 75.5 & 64.5 & 74.2 & 73.4  \\  \hline
        OKHDP\cite{li2021online} &  4-Stack HG  &  2-Stack HG &  25.5 & 72.8 & 91.5 & 79.5 & 69.9 & 77.1 & 75.6 \\
        OKHDP\cite{li2021online} &  8-Stack HG  &  4-Stack HG &  47.0 & 74.8 & 92.5 & 81.6 & 72.1 & 78.5 & 77.4 \\  \hline
        DistilPose\cite{ye2023distilpose} & HRnet-W48 & stemnet &  2.4 & 71.6 & - & - & - & - & - \\
        DistilPose\cite{ye2023distilpose} & HRnet-W48 & HRnet-W48-stage3 &  10.3 & 74.4 & - & - & - & - & - \\ \hline
        DWPose-M\cite{yang2023effective} & RTMPose-L & RTMPose-M &  1.9 & 74.9(\textcolor{red}{+0.6}) & 90.1(\textcolor{red}{+0.1}) & 81.8(\textcolor{red}{+0.5}) & 71.3(\textcolor{red}{+0.6}) & 81.7(\textcolor{red}{+0.7}) & 79.9(\textcolor{red}{+0.6})  \\ 
        DWPose-S\cite{yang2023effective} & RTMPose-L & RTMPose-S &  0.7 & 72.0(\textcolor{red}{+0.1}) & 89.7(\textcolor{red}{+0.3}) & 79.3(\textcolor{blue}{-0.1}) & 68.4(\textcolor{red}{+0.1}) & 78.5(\textcolor{red}{+0.1}) & 77.0(\textcolor{red}{+0.0})  \\ 
        DWPose-T\cite{yang2023effective} & RTMPose-L & RTMPose-T &  0.4 & 67.7(\textcolor{blue}{-0.2}) & 88.3(\textcolor{red}{+0.2}) & 75.4(\textcolor{blue}{-0.1}) & 64.5(\textcolor{red}{+0.0}) & 73.9(\textcolor{blue}{-0.3}) & 73.3(\textcolor{blue}{-0.1})  \\  \hline
        \textbf{AgentPose-M} & \textbf{RTMPose-L} & \textbf{RTMPose-M} &  \textbf{2.2} & \textbf{74.9(\textcolor{red}{+0.6})} & \textbf{90.5(\textcolor{red}{+0.5})} & \textbf{81.9(\textcolor{red}{+0.6})} & \textbf{71.4(\textcolor{red}{+0.7})} & \textbf{81.6(\textcolor{red}{+0.6})} & \textbf{79.8(\textcolor{red}{+0.5})}   \\ 
        \textbf{AgentPose-S} & \textbf{RTMPose-L} & \textbf{RTMPose-S} &  \textbf{0.9} & \textbf{72.1(\textcolor{red}{+0.2})} & \textbf{89.6(\textcolor{red}{+0.2})} & \textbf{79.3(\textcolor{blue}{-0.1})} & \textbf{68.5(\textcolor{red}{+0.2})} & \textbf{78.6(\textcolor{red}{+0.2})} & \textbf{77.2(\textcolor{red}{+0.2})}  \\ 
        \textbf{AgentPose-T} & \textbf{RTMPose-L} & \textbf{RTMPose-T} &  \textbf{0.6} & \textbf{68.3(\textcolor{red}{+0.4})} & \textbf{88.5(\textcolor{red}{+0.4})} & \textbf{76.0(\textcolor{red}{+0.5})} & \textbf{65.2(\textcolor{red}{+0.7})} & \textbf{74.3(\textcolor{red}{+0.1})} & \textbf{73.8(\textcolor{red}{+0.4})}  \\  \hline
    \end{tabular} }
    \label{Table:SOTA}
    \vspace{-1.5em}
\end{table*}

\subsubsection{Dynamic Distribution Modulation}

Dynamic distribution modulation is powered by a lightweight score-based diffusion model $S_\theta (\cdot)$, which consists of two bottleneck blocks\cite{He_2016_CVPR} and a $1 \times 1$ convolution layer. 
This module employs the reverse VP-SDE to capture the underlying distribution of teacher features and calibrate the knowledge of the student.

Regarding noisy teacher features $F_{tea}(t)$, the diffusion model learns the corresponding noise-perturbed score function $\nabla_{F_{tea}} \log p_t(F_{tea})$ to guarantee that generated features obey the distribution of teacher features in the reverse VP-SDE\cite{song2021scorebased}. Since $\nabla_{F_{tea}} \log p_t(F_{tea})$ can represent the gradient of $F_{tea}(t)$ in the data space
and guide noisy samples in the reverse VP-SDE to gradually align with the distribution of teacher feature $p_{t=0}(F_{tea})$, the training loss is expressed as follows:
\begin{align}
\mathcal{L}_{diff} \! &= \! \left\| S_\theta(F_{tea}(t), t) \! - \! \nabla_{F_{tea}} \log p_{0t}(F_{tea}(t)|F_{tea}(0)) \right\|^2 \nonumber \\
&= \left\| S_\theta(F_{tea}(t), t) \! - \! z (t)\right\|^2  \label{loss_diff_train}
\end{align}
where noisy teacher feature $F_{tea}(t)$ and corresponding timestep $t$ are used as the input of the diffusion model $S_\theta (\cdot)$.

For noisy student features $F_{stu}(t_s)$, we treat them as the denoising target $x(t)$ in the reverse VP-SDE:

\begin{equation}
    dx(t) = -\beta(t)[\frac{x}{2}+\nabla_{x(t)}\log p_t(x)]dt+ \sqrt{\beta(t)}d\bar w
    \label{reverse}
\end{equation}
where $\bar w$ has the same meaning as $w$ with opposite time course, 
noisy student feature $F_{stu}(t_s)$ and timestep $t_s$ are designated as optimization object $x(t)$ and the start step of the reverse VP-SDE.
Furthermore, since we train the diffusion model $S_\theta(\cdot)$ to fit the noise-perturbed score function $\nabla_{F_{tea}} \log p_t(F_{tea})$ of teacher feature, the solution of Eq.\ref{reverse} can be approximated with the Euler-Maruyama method\cite{bayram2018numerical} to discretize for ease of calculation:
\begin{equation}
\begin{aligned}
    x(t \! - \! \Delta t) \! &= \! \frac{1}{\sqrt{1-\beta(t)\Delta t}}(x(t) \! + \! \beta(t)\Delta t S_\theta(x(t), t))   \\
     &+\sqrt{\beta(t)\Delta t} z , z \sim N(0,I)  \label{update}
\end{aligned}
\end{equation}

\subsection{Autoencoder}

To avoid the additional computational burden brought by the feature agent, we introduce a simple linear autoencoder to reduce dimensions of features processed by the feature agent. The linear autoencoder is composed of only two $1 \times 1$ convolution layers. 

To ensure the quality of compressed features, we apply a reconstruction loss $\mathcal{L}_{Rec}$ to minimize the gap between the teacher feature $F_{tea}$ and the reconstructed feature $\Tilde{F}_{tea}$:
\begin{align}
    \mathcal{L}_{rec}=\left\|\Tilde{F}_{tea}-F_{tea}\right\|^{2} \label{9}
\end{align}
It should be noted that latent teacher features $\hat{F}_{tea}$ as training samples of the diffusion model are detached and have no gradient backward from the diffusion model. Therefore, the autoencoder is trained solely using the reconstruction loss $\mathcal{L}_{Rec}$. 

For the student feature, we also employ two $1\times1$ convolution layers to match the number of channels. The first layer adjusts the number of student features in the channel to match that of $\hat{F}_{tea}$, while the second ensures that the denoised features $\bar{F}_{stu}$ are compatible with the configuration of the detection head of the student model.

\subsection{Training and Inference}\label{section:loss}
The overall loss function is composed of the following five components: task loss $\mathcal{L}_{task}$, autoencoder's reconstruction loss $\mathcal{L}_{rec}$, diffusion loss $\mathcal{L}_{diff}$, pose feature distillation loss $\mathcal{L}_{fea}$, and logit distillation loss $\mathcal{L}_{logit}$.
\begin{align}
    \mathcal{L}_{W} \! = \! \mathcal{L}_{task} \! + \! \mathcal{L}_{rec}& \! + \! \mathcal{L}_{diff} \! + \! R(E)(\mathcal{L}_{fea} \! + \! \mathcal{L}_{logit}) \label{10} \\
    R(E) &= 1-(E-1)/E_{max}  \label{target-decay}
\end{align}
where weight decay function $R(E)$ is determined by current epoch $E$ and maximum epoch $E_{max}$. 
This function facilitates the student model and feature agent in focusing on leveraging the teacher's knowledge to address tasks more effectively, rather than blindly imitating the teacher's output.

Since we adopt RTMPose\cite{jiang2023rtmpose} as the basic estimation network, we can utilize the task loss of RTMPose as $\mathcal{L}_{task}$ for our framework:
 \begin{align}
     \mathcal{L}_{task}=-\frac{1}{NKL}\sum_{n=1}^{N}\sum_{k=1}^{K}W_{n,k}\sum_{l=1}^{L}V_{l}\log(P_{stu}^{l}) \label{11}
 \end{align}
where $W_{n,k}$ is the target weight mask, $V_{l}$ is the label value, and $P_{stu}^{l}$ is the student's prediction. $N, K, L$ represent the batch size, the number of keypoints, and the length of the $x$ or $y$ localization bins respectively.

As for the logit distillation loss $\mathcal{L}_{logit}$, we adopt same form as $\mathcal{L}_{task}$, but drop the target weight mask $W_{n,k}$:
\begin{align}
     \mathcal{L}_{logit}=-\frac{1}{NKL}&\sum_{n=1}^{N}\sum_{k=1}^{K}\sum_{l=1}^{L}P_{tea}^{l}\log(P_{stu}^{l}) \label{12} 
 \end{align}
where $P_{tea}^{l}$ is teacher prediction. Simultaneously, we take the mean square error between $\hat{F}_{tea}$ and $\bar{F}_{stu}$ as the feature distillation loss:
\begin{align}
    \mathcal{L}_{fea}=\frac{1}{CHW}\left\|\hat{F}_{tea}-\bar{F}_{stu}\right\|^2 \label{13}
\end{align}
among them, $H$, $W$, and $C$ respectively denote the height, width, and channel of features.

During inference, the student backbone first extracts features from the input images and uses a $1\times1$ convolution layer to adjust the number of feature channels. Subsequently, we add a suitable amount of Gaussian noise to features in forward VP-SDE, where the timestep $t_s$ is set to $0.4$. Then, the feature agent progressively removes the noise from noisy features. Additionally, we use another $1\times1$ convolution layer to ensure denoised features meet the parameter settings of the student detection head. Finally, the student model outputs predictions about the human pose.

\section{Experiment}

\subsection{Dataset}
COCO\cite{lin2014microsoft} is a widely used 2D human pose estimation dataset that contains 200K image data with different poses. Additionally, it provides 17 keypoints annotation for each human instance and is split into train2017 and val2017. 

\subsection{Implementation details} 
We utilize RTMPose\cite{jiang2023rtmpose} as our foundational pose estimation network and DWPose\cite{yang2023effective} as our benchmark framework. Therefore, AgentPose and DWPose utilize RTMPose-L as the teacher and other smaller RTMPose versions as students.
For all experiments, AdamW\cite{loshchilov2017decoupled} serves as the optimizer.  
The student feature distribution perturbation coefficient
$t_s$ is set to $0.4$, and other hyperparameters based on default configurations of MMPose\cite{mmpose2020}. The size of input image is set to $256 \times 192$.

\subsection{Method comparison}
\noindent \textbf{Comparison with state-of-the-art methods.}
As shown in Tab.\ref{Table:SOTA}, we compared the proposed method with SOTA methods\cite{ye2023distilpose, yang2023effective, li2021online, jiang2023rtmpose} and the results demonstrate that AgentPose achieves top-tier performance. 
Specifically,
AgentPose-M, achieves 74.9 AP with a mere 2.2 GFLOPs.
In comparison, alternative methods necessitate a substantially greater computational expenditure to match this level of performance.
Moreover, the performance degradation exhibited by DWPose-T underscores the difficulties posed by the capacity gap challenge, which AgentPose adeptly addresses with only an additional 0.2 GFLOPs.

\noindent \textbf{Comparison with other pose distillation methods.}
We have conducted a comparison of AgentPose against other pose distillation techniques that tackle the capacity gap curse, as cited in the literature \cite{mirzadeh2020improved, zhu2022teach, son2021densely, li2022shadow, li2021reskd, ding2023skdbert, rao2023parameter, wang2023bridging, qian2022switchable}. Tab.\ref{Table:various_kd} clearly demonstrates that AgentPose is on par with other methods in boosting the performance of student models, and it does so with a negligible increase in computational resources. Although there are methods that may marginally surpass AgentPose in terms of accuracy, they generally require intricate training procedures or higher computational costs.

\begin{table}[ht]
\vspace{-0em}
    \centering
    \caption{Comparison with other knowledge distillations to evaluate the robustness to the capacity gap curse.}
    \resizebox{0.9\linewidth}{!}{
  \begin{tabular}{|c|c|c c c c|}
    \hline
        \textbf{Method} & \textbf{GFLOPs}  & $\bm{AP}$ & $\bm{AP^{50}}$ & $\bm{AP^{75}}$ & $\bm{\bm{AR}}$ \\ \hline 
        TAKD\cite{mirzadeh2020improved}  & 0.36 &  66.6 & 87.7 & 74.3 & 72.2  \\  
        TLLM\cite{zhu2022teach} & 0.36 &   68.1 & 88.2 & 76.0 & 73.6  \\ 
        DGKD\cite{son2021densely} & 0.36 &  68.1 & 88.1 & 75.8 & 73.5\\ 
        SKD\cite{ding2023skdbert} & 0.36 &  68.2 & 88.2 & 75.7 & 73.5  \\  
        PESF-KD\cite{rao2023parameter} & 0.36 &  62.8 & 85.3 & 69.6 & 69.7   \\ 
        ResKD\cite{li2021reskd} & 0.72 &  68.6 & 88.5 & 76.1 & 73.9  \\ 
        SHAKE\cite{li2022shadow} & 0.36 &  63.0 & 85.9 & 70.2 & 69.3  \\
        ABML\cite{wang2023bridging} & 0.36 & 64.1 & 86.3 & 71.6 & 70.1\\
        SwitKD\cite{qian2022switchable} & 0.36 & 64.5 & 86.4 & 71.2 & 70.4  \\ \hline
        \textbf{AgentPose} & \textbf{0.59} & \textbf{68.3} & \textbf{88.5} & \textbf{76.0} & \textbf{73.8}\\ \hline
    \end{tabular}}
    \label{Table:various_kd}
    \vspace{-1em}
\end{table}

\subsection{Ablation study}
\subsubsection{Effect of starting timestep}
 
The starting timestep plays a pivotal role in determining the intensity of noise perturbations on the student feature, significantly affecting the trajectory of reverse VP-SDE on the student feature and the model accuracy.
As shown in Tab. \ref{tab:start_timestep}, when the starting timestep equals 0.4, it achieves the best accuracy. 
The rationale behind this is that when the starting timestep is set too small, the noise samples retain a significant amount of student pose information that cannot match the teacher's features, preventing the feature agent from further supplementing and optimizing the noise samples. Conversely, when the starting timestep is too large, the semantic information of noise samples is severely compromised. As a result, the features randomly generated by the feature agent may not correspond to the content of the original images. 

\begin{table}[ht]

    \caption{Effect of starting timestep. }
    \centering
    \resizebox{0.80\linewidth}{!}{
    \begin{tabular}{|c|c|c |c|}
    \hline
       \textbf{Method}  &  \textbf{Starting timestep} &  \textbf{AP} & \textbf{AR} \\ \hline
        \multirow{4}{*}{AgentPose-T} &  0.2  & 68.32 & 73.71 \\
          & 0.4 & \textbf{68.33} & 73.75 \\
          & 0.6 & 68.26 & \textbf{73.92} \\ 
          & 0.8 & 68.11 & 73.63 \\ \hline
    \end{tabular}
    } 
    \label{tab:start_timestep}
    \vspace{-1em}
\end{table}

\subsubsection{Effect of latent feature dimension in autoencoder}
As shown in Table \ref{tab:autoencoder}, the dimension of latent feature in the autoencoder significantly influences AgentPose. It is clear that the autoencoder plays an effective role in reducing the model's computational complexity. When the dimension of the latent feature is set to 128 or 256, the computational overhead is markedly decreased, amounting to approximately 30\% of its original cost. Nevertheless, achieving such a significant reduction in computational overhead may come with a trade-off in performance.

\begin{table}[ht]
\vspace{0em}
    \caption{Effect of latent feature dimension in autoencoder.}
    \centering
    \resizebox{0.8\linewidth}{!}{
    \begin{tabular}{|c|c|c| c| c |c|}
    \hline
          ~  & \textbf{w/o AE}  &\textbf{128} & \textbf{256} & \textbf{512} & \textbf{1024}  \\
    \hline
         AP & 67.92 & 68.16 & 68.27 & 68.33 & 68.36                 \\
        GFLOPs & 1.24 & 0.38 & 0.42 & 0.59 & 1.24                         \\
    \hline
    \end{tabular} }
    \label{tab:autoencoder}
\end{table}

\subsubsection{Effect of inference steps}
We evaluate the performance of AgentPose at various inference steps, as depicted in Tab.\ref{tab: inference step}. The model achieves its peak accuracy of 68.33 AP with 5 inference steps. Beyond this point, adding more inference steps leads to a decrease in performance and slows down the inference speed.

\begin{table}[ht]
\vspace{-0.5em}
    \caption{Results of AgentPose with different inference steps.}
    \centering
    \resizebox{0.6\linewidth}{!}{
    \begin{tabular}{|c|c| c|}
    \hline
        \textbf{Inference step} & \textbf{AP} & \textbf{GFLOPs} \\ \hline
        1 &  68.19 & 0.43 \\ 
        3 & 68.14 & 0.51 \\ 
        5 & 68.33 & 0.59 \\ 
        10 &  68.3 & 0.79 \\ 
        15 & 68.26 & 0.98 \\ \hline
    \end{tabular} }
    \label{tab: inference step}
    \vspace{-1em}
\end{table}

\section{Conclusion}

This paper proposed AgentPose, an innovative pose distillation framework for human pose estimation. We narrow the distribution discrepancy between teacher and student features by employing a feature agent. Notably, the feature agent harnesses noisy student features within the reverse VP-SDE process, enabling automatic adjustments to balance feature quality and content consistency, and thus delivering compatible features to the student model.
Comprehensive experimental results further demonstrate that AgentPose adeptly calibrates the student knowledge, enhancing the consistency between the student and teacher models, and facilitating the transfer of the teacher's knowledge.

\section*{Acknowledgment}

This work is supported by the National Natural Science Foundation of China (Grant No. 62202241), Jiangsu Province Natural Science Foundation for Young Scholars (Grant No. BK20210586), NUPTSF (Grant No. NY221018) and Double-Innovation Doctor Program under Grant JSSCBS20220657.

\bibliographystyle{IEEEbib}
\bibliography{main}

\end{document}